\documentclass[letterpaper, 10 pt, conference]{ieeeconf}  % Comment this line out if you need a4paper

\IEEEoverridecommandlockouts                              % This command is only needed if 
\usepackage{graphics} % for pdf, bitmapped graphics files
\usepackage{epsfig} % for postscript graphics files
\usepackage{mathptmx} % assumes new font selection scheme installed
\usepackage{times} % assumes new font selection scheme installed
\usepackage{amsmath} % assumes amsmath package installed
\usepackage{amssymb}  % assumes amsmath package installed
\usepackage{graphicx}
\usepackage{caption}
\usepackage{mathrsfs}
\usepackage[ruled,vlined]{algorithm2e}
\usepackage[font={small}]{caption}

\usepackage{balance}   %ADDED 20220224 HES

\usepackage[utf8]{inputenc}
\usepackage[english]{babel}
\usepackage[backend=biber,url=false,sorting=none,citestyle=numeric-comp]{biblatex}
\newcommand{\stack}[2]{\begin{tabular} {@{}c@{}}#1 \\ #2 
\end{tabular}}

\title{\LARGE \bf
Humanoid Path Planning over Rough Terrain using Traversability Assessment
}

\author{Stephen McCrory$^{1,2}$, Bhavyansh Mishra$^{1,2}$, Jaehoon An$^{1,4}$, Robert Griffin$^{1,2}$, \\ Jerry Pratt$^{1,2,3}$ and Hakki Erhan Sevil$^{2}$% <-this % stops a space
\thanks{$^{1}$Author is with the Institute of Human and Machine Cognition (IHMC),
        40 S Alcaniz St, Pensacola, FL 32502, USA
        {\tt\small author@ihmc.org}}%
\thanks{$^{2}$Author is with the University of West Florida (UWF),
        11000 University Pkwy, Pensacola, FL 32514, USA
        {\tt\small author@uwf.edu}}%
\thanks{$^{3}$Author is with Boardwalk Robotics, Inc., 11 Gathering Green E, Pensacola, FL 32502, USA {\tt\small author@boardwalkrobotics.com}}
\thanks{$^{4}$Author is with Pusan National University, Busan, South Korea {\tt\small author@pusan.ac.kr}}
\thanks{*This work was supported through ONR Grant No. N00014-19-1-2023 and NASA Grant No. 80NSSC20M0197.}% <-this % stops a space
}

\begin{document}
\maketitle
\thispagestyle{empty}
\pagestyle{empty}
%%%%%%%%%%%%%%%%%%%%%%%%%%%%%%%%%%%%%%%%%%%%%%%%%%%%%%%%%%%%%%%%%%%%%%%%%%%%%%%%
\begin{abstract}

We present a planning framework designed for humanoid navigation over challenging terrain. This framework is designed to plan a traversable, smooth, and collision-free path using a 2.5D height map. The planner is comprised of two stages. The first stage consists of an A* planner which reasons about traversability using terrain features. A novel cost function is presented which encodes the bipedal gait directly into the graph structure, enabling natural paths that are robust to small gaps in traversability. The second stage is an optimization framework which smooths the path while further improving traversability. The planner is tested on a variety of terrains in simulation and is combined with a footstep planner and balance controller to create an integrated navigation framework, which is demonstrated on a DRC Boston Dynamics Atlas robot.

\end{abstract}
%%%%%%%%%%%%%%%%%%%%%%%%%%%%%%%%%%%%%%%%%%%%%%%%%%%%%%%%%%%%%%%%%%%%%%%%%%%%%%%%
\section{INTRODUCTION}

% Problem outline/motivation and high level discussion of challenges to solving it
Legged robots have increasingly become capable of robust locomotion and navigating over unstructured terrain. A major advantage of legged locomotion is an ability to traverse terrain which contains obstacles, gaps or other challenges intractable to wheeled or tracked platforms. For many real-world applications of legged platforms, it is a requirement that such terrain can be navigated with a high degree of autonomy. Path planning for legged robots requires reasoning about the platform's capabilities and can be difficult to deploy when combined with practical limitations such as sensor noise. While there has been significant work towards this goal, finding a robust and operational solution to humanoid path planning remains an open problem.

% Terrain cost models and shortcomings
This paper presents a navigation path planner designed to enable humanoid locomotion over rough terrain and is intended to be used as a heuristic for a lower-level footstep planner. We build on the approach of many existing planners and perform a sample-based graph search which includes a traversability cost model \cite{brunner2013hierarchical, wermelinger2016navigation, klamt2017anytime}. The main contribution of our formulation is setting up the planner to use these traversability costs in a way that reflects the bipedal gait. Additionally, we include checks to prevent cutting corners, maintain a safe distance from obstacles and find reliable routes for ascending and descending terrain.

\begin{figure}[t]
\centering
  \includegraphics[width=\columnwidth]{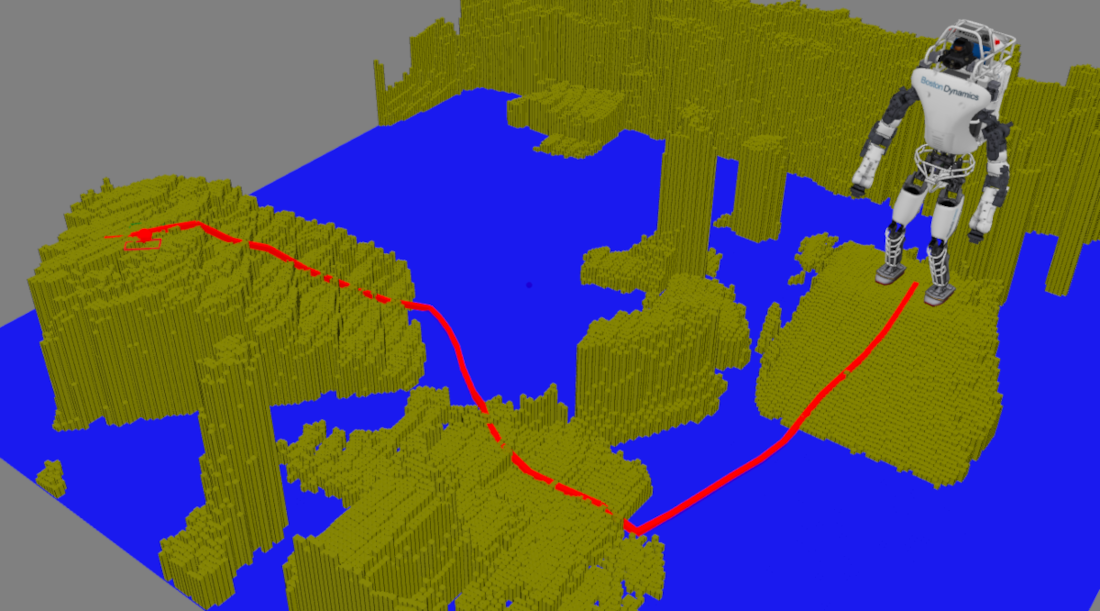}
  \caption{Path planned over an obstacle course consisting of a ramp, cinder blocks, and a staircase. A 2.5D height map, the robot's initial configuration and an operator-specified goal is given to the planner as input.}
  \label{fig:intro_figure}
\end{figure}

% Approach
Our approach first performs an A* search over a 2.5D height map by sampling terrain in the vicinity of each node to measure traversability. This initial path is then optimized using gradient descent in order to smooth the path while further improving its quality. We select A* for a few reasons, the first being that our cost functions are sample-based and preclude the use of a closed-form solver. The second is for the practical reason that the A* planning process lends itself to logging and visualization more than randomized or probabilistic graph-search approaches. The planner is tested extensively using real sensor data on a variety of terrains containing stairs, stepping stones, ramps, cinder blocks and large obstacles (Fig. \ref{fig:intro_figure}). We also share results from testing on a DRC Atlas robot by integrating with an existing balance controller and footstep planner.

\section{RELATED WORK}

% Two stage approach
A common approach in path planning for mobile robots is to first plan an approximate body path which is then used as a heuristic for a low-level planner. Chestnutt \cite{chestnutt2007navigation} developed one of the first such formulations for a humanoid in which a collision-free body path is computed using a floor plan. This path is then used as a heuristic for an A* footstep search.

% Reasoning about traversability
In developing more capable path planners, work has been done to create heuristic cost models for terrain traversability. Metrics such as local terrain slope, roughness and curvature are commonly used \cite{winkler2015planning}. During planning, these metrics can be used in combination with a geometric model of the robot to predict the likelihood that a node contains secure footholds \cite{wermelinger2016navigation}. If sufficiently accurate sensing is available terrain can be modelled using planar regions \cite{fallon2015, mishra2021gpu} or curved surfaces \cite{kanoulas2017vision}. While this can increase the accuracy of the terrain cost model, the sensor update rate is usually either too slow for operational applications or has a field of view that is too small for navigation planning. Learning-based terrain cost models have also been introduced \cite{garcia2018}, however this requires an extensive training pipeline with a large set of representative data. Results have also been shown for using terrain traversability as a cost in an RRT* search where a control-Lyapunov function both interpolates between nodes and provides a reactive feedback law \cite{Huang2021}.

% Path optimization
In sample-based approaches to path planning, it can be beneficial to post-process the plan. This uses the assumption that the initial planner succeeds in being in the neighborhood of an executable plan. Dolgov et al. \cite{dolgov2010path} utilized this method during the DARPA Urban Challenge by planning the path for a vehicle with hybrid A* then smoothing it with conjugate-gradient descent. While this approach was implemented for a wheeled vehicle, it readily extends to legged platforms. Other methods for path improvement include waypoint pruning and increasing obstacle clearance when possible \cite{dang2020graph}.

% Model-based approaches
Many model-based approaches have been formulated which generate stable trajectories over long horizons. Fernbach et al. \cite{fernbach2017kinodynamic} plan a path using RRT-connect \cite{kuffner2000rrt}, checking for collisions with a simple bounding volume and available contacts by approximating reachability. Surface normals are used to model contact force constraints and a linear program is used to determine acceleration bounds between nodes. A similar approach in \cite{norby2020fast} also uses RRT-connect but performs stability checks by randomly sampling feasible ground reaction forces, yielding fast planning durations on the order of seconds.

% Similarity of approach
Our approach is most similar to \cite{brunner2013hierarchical} and \cite{wermelinger2016navigation}. Though these works perform RRT* and not an A* search, the core of the planner is reasoning about traversability without modelling exact contacts. We build on these approaches with a graph structure and novel cost framework targeted at bipedal navigation.
\section{A* BODY PATH PLANNING}

Our planning process begins by performing an A* search to find collision-free and traversable waypoints. We model the environment using a 2.5D height map which has a cell discretization of $\Delta_h$ (see Sec. \ref{subsec:height_map} for details on mapping). This height map is provided as input to the planner along with the robot's start position $x_0$ and a goal position $x_G$. The output is a sequence of waypoint positions $x_0, \ldots x_N=x_G$. This section describes the graph and cost framework used for the A* search and the following section describes an optimization framework for smoothing this initial path.

\subsection{Graph Structure}

\begin{figure}[t]
  \vspace{2mm}
\centering
  \includegraphics[width=4.4cm]{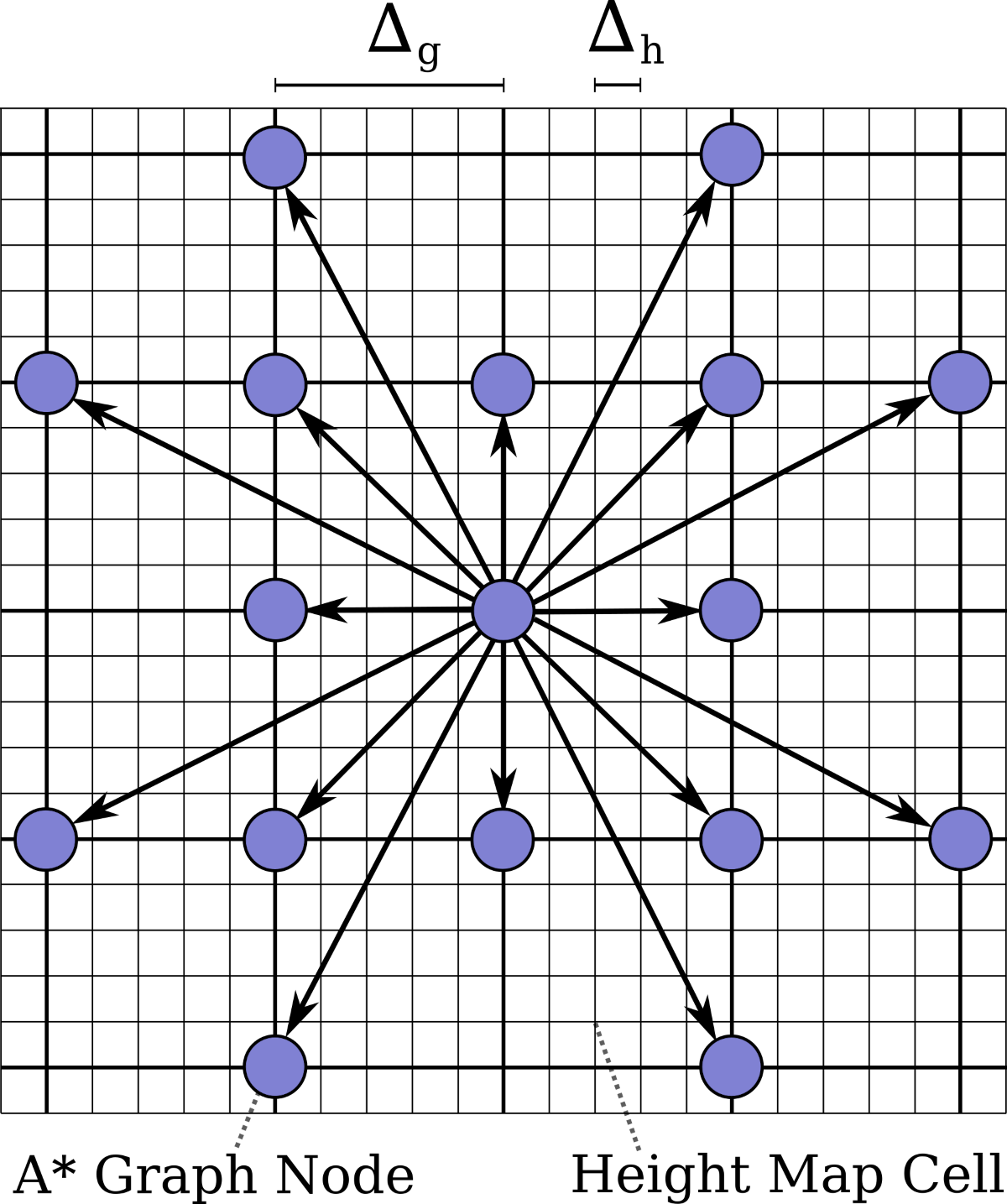}
  \caption{The A* graph plans along a grid with discretization $\Delta_g$ and uses a 16-connected transition model. This shows an example node expansion superimposed on the height map grid, which has discretization $\Delta_h$.}
  \label{fig:graph}
\end{figure}

The A* planner samples nodes in a grid with discretization $\Delta_g$, which is set to a multiple of the height map discretization $\Delta_h$. A \textit{graph node} represents an unoriented robot position. It is defined by a pair of integer coordinates $n_i=(x_i, y_i)$ which maps to the continuous coordinates $\textbf{x}_i=(\Delta_g x_i, \Delta_g y_i)$. A \textit{graph edge} represents an oriented transition between positions, defined as a pair of nodes $(n_i, n_j)$. Unless otherwise stated, node $i$ and $j$ represent the parent and child nodes of a generic edge, respectively. The transition model expands to 16 neighbor nodes, as depicted in Fig. \ref{fig:graph}.

Each node is assigned a height based on sampling the terrain within a fixed radius of the node. The node height is the mean of sampled heights within $[h_m-\epsilon,h_m]$, where $h_m$ is the maximum sampled height and $\epsilon$ is a fixed window. An \textit{edge frame} $\mathscr{E}_{ij}$ is defined at node $j$ where $\hat{\textbf{z}}_{ij}$ is parallel with the world-frame $z$-axis, $\hat{\textbf{x}}_{ij}$ is parallel with $\textbf{x}_j - \textbf{x}_i$ and $\hat{\textbf{y}}_{ij} = \hat{\textbf{z}}_{ij} \times \hat{\textbf{x}}_{ij}$, as shown in Fig. \ref{fig:astar_costs}a.

\subsection{Edge Feasibility}

Four feasibility checks are performed on a candidate edge and edges that do not meet any of these criteria are not added to the A* graph. First, the parent and child nodes must be within the $xy$ domain of the height map. Second, a maximum incline magnitude between the two nodes is imposed, where the incline of an edge is given by
\begin{equation}
    \theta_{ij} = \tan^{-1}{\frac{z_j - z_i}{ | \textbf{x}_j - \textbf{x}_i | }}.
\end{equation}

Third, a collision check is performed by approximating the bounding volume of the robot as a box that is aligned with and vertically offset from the edge frame. Fig. \ref{fig:astar_costs}b depicts this bounding volume and resulting sampled cells. If the height map has any intersection with the bounding volume, the graph edge is infeasible. The fourth check is based on the edge's traversability, as described in the next subsection.

\subsection{Terrain Traversability} \label{subsec:as_trav}

\begin{figure}[t]
  \vspace{2mm}
\centering
  \includegraphics[width=\columnwidth]{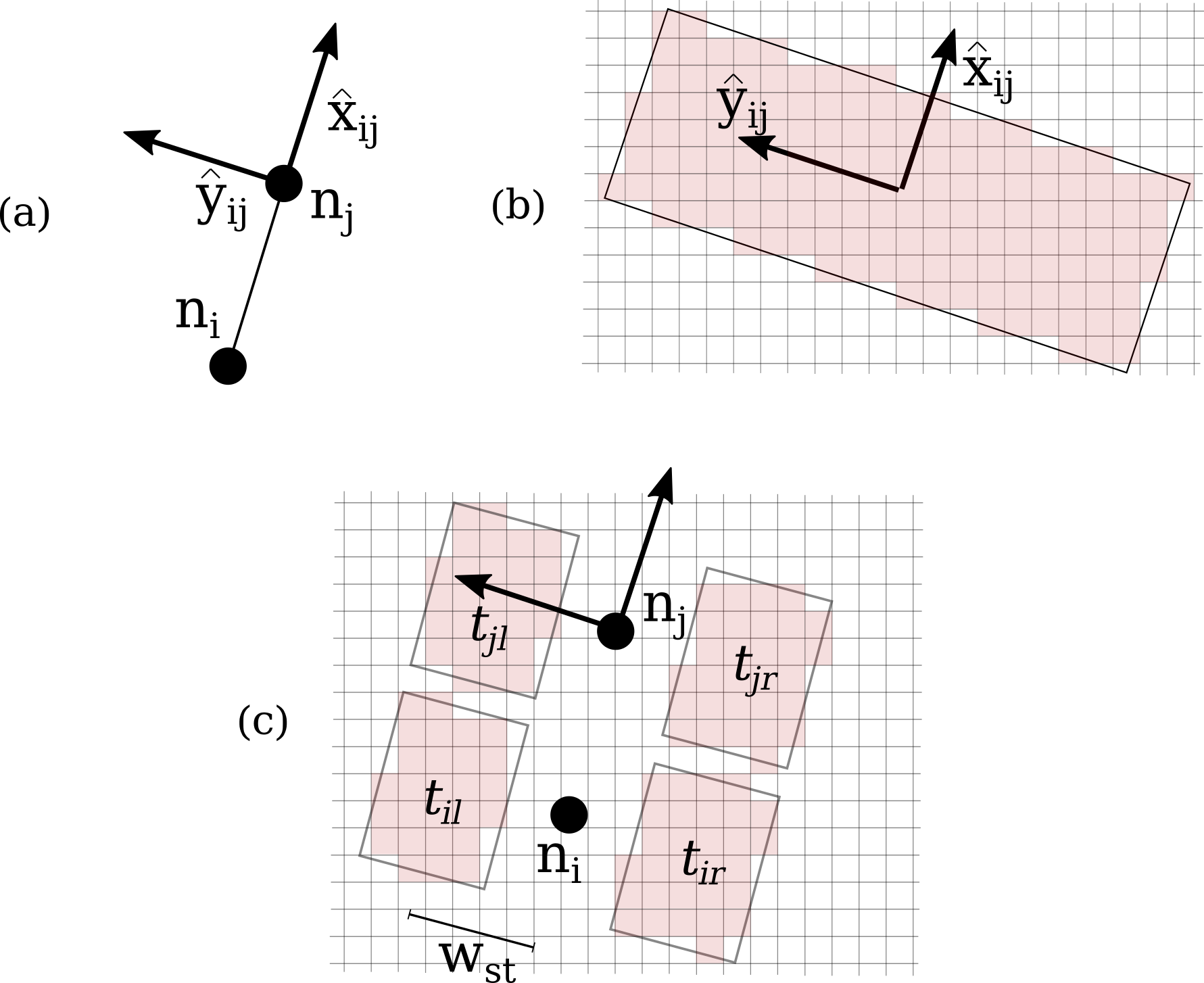}
  \caption{A graph edge's geometry is used to model the robot's configuration by defining an edge frame (a). The robot's bounding volume is expressed in this frame as a box and used to check for collisions (b). Regions of the height map on both sides of the parent and child nodes are sampled to reason about the edge's traversability based on the robot's preferred stance width $w_{st}$ (c).}
  \label{fig:astar_costs}
  \vspace{-4mm}
\end{figure}

This section describes a cost framework penalizing nodes with low traversability while being robust to sufficiently small patches of non-traversable terrain. First a nominal stance width $w_{st}$ is defined which represents the preferred distance of the robot's foot to the sagittal plane during locomotion. Using $w_{st}$ and the edge frame, we compute the nominal foothold positions of the parent and child nodes. We then sample the height map in a rectangular region around each nominal foothold, as shown in Fig. \ref{fig:astar_costs}c. Each region is assigned a traversability score $t \in [0, 1]$, with 1 representing full traversability. This score is computed as the percentage of sampled height map cells with a sufficiently low incline and within a certain proximity of the node's height. The surface normal is computed using RANSAC, following the approach in \cite{marion2016}.

Given the four nominal traversability scores we define two traversability metrics. An edge's \textit{foothold traversability}, which represents the availability of a single foothold, is given by 
\begin{equation}
    t_f = \max (t_{jl}, t_{jr}),
\end{equation} where $t_{ks}$ is the score of the region on side $s$ of node $k$. Similarly, an edge's \textit{stance traversability} represents the availability of a pair of footholds between the two nodes,
\begin{equation}
    t_s = \max (\sqrt{t_{jl} t_{ir}}, \sqrt{t_{jr} t_{il}}).
\end{equation}

A minimum threshold $t_{f,min}$ is defined for foothold traversability and is used as a feasibility check.

\subsection{Contour Cost} \label{subsec:contour_cost}

We find that often the most reliable strategy for changing elevation is the path of (locally) steepest ascent/descent and that trying to achieve lateral progress while changing elevation is undesirable. In our experience, this heuristic increases reliability for stairs, ramps, and stepping stones. We encode this undesirable behavior as graph edges where the local terrain has significant roll and pitch. In other words, this heuristic asserts it is better to walk parallel or perpendicular to contour lines but not in between, hence name \textit{contour cost} is used.

A cost term is included to encourage paths which follow this heuristic. The surface normal $\hat{\textbf{n}}_j$ is first computed using a least-squares fit in the neighborhood of the child node. A least-squares fit is well suited for modelling terrain contours, as opposed to RANSAC which could give a uniformly vertical normal on stairs, for example. The contour cost is the product of the edge's incline (pitch) and the roll of $\hat{\textbf{n}}_{j}$ expressed in the edge frame:
\begin{equation} \label{eq:astar_contour_cost}
    c_c = | \theta_{ij} \sin ^{-1} (\hat{\textbf{n}}_{j} \cdot \hat{\textbf{y}}_{ij}) |.
\end{equation}

\subsection{A* Algorithm}
The edge cost function $g_{ij}$ is given in Eq. \ref{eq:complete_cost}, where the terms $w_f$, $w_s$ and $w_c$ are weight factors for costs based on foothold traversability, stance traversability and the contour heuristic (respectively).
\begin{equation} \label{eq:complete_cost}
  \begin{aligned}
g_{ij} = |\textbf{x}_j - \textbf{x}_i| + w_f (1 - t_f) + w_s (1 - t_s) + w_c c_c
 \end{aligned}
\end{equation}

The A* planner uses the standard heuristic function $h_i = |\textbf{x}_i - \textbf{x}_G|$ which is the candidate node's distance to the goal. Each iteration, the node with the lowest total cost $g_i + h_i$ is expanded using the sixteen-connected transition model, where $g_i$ is the least-cost path from to node $i$. Candidate edges are then checked for feasibility and valid edges are assigned a cost and added to a priority queue. The planner continues until the goal is reached, the queue is empty or a timeout is reached.
\section{WAYPOINT OPTIMIZATION}

The primary goal of the A* planner is to find rough yet feasible paths which can be subsequently processed by a local optimizer. Sampling a sparse grid is inherently limited by the grid resolution and will result in non-smooth paths. Therefore we implement a path optimizer for both smoothing the path while utilizing the same traversability metrics used for planning. We employ a similar approach to \cite{dolgov2010path}, in which conjugate-gradient descent is used for smoothing and obstacle clearance. To do this, we formulate versions of the A* traversability and contour costs (Sec. \ref{subsec:as_trav},D) in continuous space. Since these costs require sampling the height map, they prohibit the use of conjugate-gradients, and we instead perform a standard gradient descent. We define the quantities $\Delta \textbf{x}_i = \textbf{x}_i - \textbf{x}_{i - 1}$ and $\Delta \phi_i = | \tan ^{-1} \frac{\Delta y_{i+1}}{\Delta x_{i+1}} - \tan ^{-1} \frac{\Delta y_i}{\Delta x_i} | $. The cost function is given in Eq. \ref{eq:opt_costs}, containing terms to reward \textit{uniform spacing}, \textit{smoothness}, \textit{obstacle clearance}, \textit{traversability} and following the \textit{contour heuristic}. This function is constructed using distinct $\sigma$ functions, which are defined through the remainder of this section.

\begin{equation} \label{eq:opt_costs}
  \begin{aligned}
c = \sum_i {w}'_u (\Delta \textbf{x}_{i+1} - \Delta \textbf{x}_i)^2 + w'_s \sigma_{s} (\Delta \phi_i - \kappa_{max}) \\
+ w'_o \sigma_o(\textbf{x}_i) + w'_t \sigma_t (\textbf{x}_i) + w'_c \sigma_c(\textbf{x}_i)
 \end{aligned}
\end{equation}

The first two terms smooth the path by rewarding uniform waypoint spacing and smooth curvature. The remaining terms are continuous-space versions of the obstacle collision check and traversability/contour costs used in the A* search. The terms $w'_u$, $w'_s$, $w'_o$, $w'_t$ and $w'_c$ are weight factors.

The gradient of the uniform-spacing term is readily computed as a linear combination of waypoint coordinates. The smoothness cost $\sigma_s$ is 0 for negative arguments and an exponential for positive arguments, in which the exponent $k$ is a positive tunable parameter. This creates a deadband so that small amounts of curvature are not penalized. Therefore the gradient is 0 when $\Delta \phi_i < \kappa_{max}$ and otherwise
\begin{equation}
   \frac{\partial \sigma_{s,i}}{\partial \textbf{x}_i} = k (\Delta \phi_i - \kappa_{max} )^{k-1} \frac{\partial \Delta \phi_i}{\partial \textbf{x}_i}.
\end{equation} 
Similar expressions for the terms $\frac{\partial \sigma_{s,i+1}}{\partial \textbf{x}_i}$ and $\frac{\partial \sigma_{s,i-1}}{\partial \textbf{x}_i}$ are included, and a full analytical expression is readily computed \cite{dolgov2010path}.

\begin{figure}[t]
  \vspace{2mm}
\centering
  \includegraphics[width=\columnwidth]{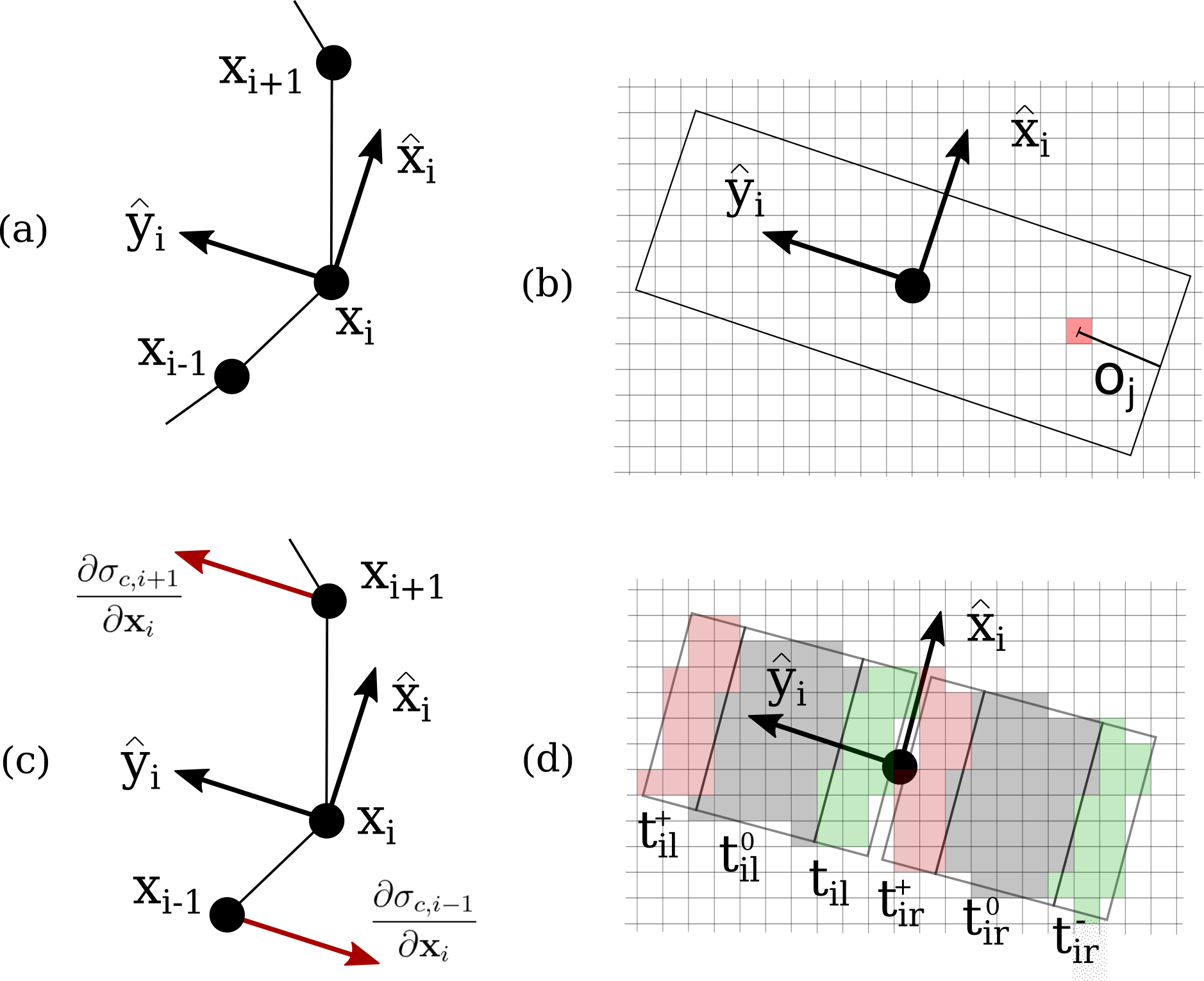}
  \caption{Waypoint optimization is performed using a local frame, shown in (a). This frame is used to check a bounding volume of the robot for collisions (b), compute the contour cost gradient (c) and sample traversable cells (d).}
  \label{fig:waypoint_opt}
\end{figure} 
The collision, traversability and contour costs are based on the robot's position and heading. Similar to the A* edge frame, we define an \textit{optimized waypoint frame} $\mathcal{O}_i$ in which the axis $\hat{\textbf{z}}_i$ is parallel to the world-frame $z$-axis, $\hat{\textbf{x}}_i$ is parallel to the vector $\textbf{x}_{i+1} - \textbf{x}_{i-1}$ and $\hat{\textbf{y}}_i = \hat{\textbf{z}}_i \times \hat{\textbf{x}}_i$ as shown in Fig. \ref{fig:waypoint_opt}a.

To compute the obstacle cost $\sigma_o$, all height map cells intersecting the robot's bounding volume are computed using the rectangular bounding volume, as shown in Fig. \ref{fig:waypoint_opt}b. For each intersecting grid cell $j$ the distance $o_j$ between the cell and the closest lateral edge is computed. When there is a non-zero number of collisions $n$, the obstacle cost is

\begin{equation}
\sigma_{o,i} = \frac{1}{n} \sum_j o^2_j.
\end{equation} This results in a simple linear gradient which is parallel to $\hat{\textbf{y}}_i$. 

As in A*, the local optimizer has a traversability cost based on the height and incline of the local terrain. We sample terrain within a nominal stepping region on either side of a waypoint and compute a corresponding traversability score as described in Sec. \ref{subsec:as_trav}. This traversability score is given as $t^0_{is} \in [0,1]$, based on sampling waypoint $i$ on side $s$. To model the gradient of this score, both lateral sides of a region are also scored, given as $t^+_{is}$ and $t^-_{is}$ (Fig. \ref{fig:waypoint_opt}d). The traversability score gradient is approximated as $t^+_{is} - t^-_{is}$, which represents how to shift a waypoint $i$ along $\hat{\textbf{y}}_i$ to increase foothold availability. However, we only want to apply this gradient to waypoints which lack traversability on the given side. To determine the extent that waypoint $i$ should have this gradient applied, a preview window $p=\{i, i\pm1, \ldots\ i\pm n_p\}$ along side $s$ is used. The maximum traversability score in this window informs whether the waypoint should be shifted, such that waypoints with a low preview score will have the gradient applied. The traversability cost gradient is defined in Eq. \ref{eq:trav_grad}. Note the sign comes from the inverse relationship of cost and traversability.
\begin{equation}\label{eq:trav_grad}
    \frac{\partial \sigma_{t,i}}{\partial \textbf{x}_i} = -\sum_{s=l,r} (1 - \max_p{t^0_{ps}}) (t^{+}_{is} - t^{-}_{is})
\end{equation}

Each waypoint has a contour cost $\sigma_c$ similar to Eq. \ref{eq:astar_contour_cost}, which we formulate in continuous space by directly modelling the gradient as a local adjustment term in Eq. \ref{eq:contour_fb}. While this corresponds to a quadratic cost term, we use both the previous and subsequent waypoints to compute the incline $\theta_i$, as opposed to the A* cost which only uses the previous. This provides a more reliable incline measurement and enables a higher cost. This cost term tries to align $\hat{\textbf{x}}_i$ perpendicular to the local contour by shifting the adjacent nodes (Fig. \ref{fig:waypoint_opt}c).
\begin{equation} \label{eq:contour_fb}
    \frac{\partial \sigma_{c,i+1}}{\partial \textbf{x}_i} = -\frac{\partial \sigma_{c,i-1}}{\partial \textbf{x}_i} = |\theta_i| (\hat{\textbf{y}}_i \cdot \hat{\textbf{n}}_i) \hat{\textbf{y}}_i
\end{equation}

\subsection{Turn Points}

When having to walk around a non-traversable area it is often preferable to turn in place instead of gradually turning over multiple steps. To promote this behavior we designate certain waypoints as \textit{turn points}, in which the curvature cost is set to zero. To determine a path's turn points we first perform a fixed number of iterations without any turn points. Then all waypoints are added to a queue and sorted in descending order of curvature cost. The head of the queue is iteratively removed and if both its curvature exceeds a threshold and position is sufficiently far from another turn point, it is designated a turn point. The optimizer then continues iterating and does not include a curvature cost for turn points.

\subsection{Gradient Descent Optimization}
\begin{table}
  \vspace{2mm}
\caption{Waypoint Optimization Weights}

\begin{center}
\begin{tabular}{|c c c|} 
 \hline
 Parameter & Description & Value \\
 \hline
 $w'_u$ & Uniform-spacing & 2 \\ 
 $w'_s$ & Smoothness & 0.7 \\ 
 $w'_o$ & Obstacle clearance & 700 \\ 
 $w'_t$ & Traversability & 20 \\ 
 $w'_c$ & Contour & 20 \\ 

 \hline
\end{tabular}
  \vspace{-3mm}
\end{center}

\label{tab:opt_params}
\end{table}

The gradient descent optimizer updates waypoint positions each optimization cycle according to Eq. \ref{eq:grad_desc_opt}, where $\gamma$ is the optimization gain and $\textbf{x}[\cdot]$ is a stacked vector of waypoint coordinates.
\begin{equation} \label{eq:grad_desc_opt}
     \textbf{x}[n+1] = \textbf{x}[n] - \gamma \nabla c
\end{equation}
After each optimization cycle, the optimized waypoint frame is recomputed. The optimizer terminates when either a maximum number of iterations is reached or the normalized (by number of waypoints) gradient magnitude is below a certain threshold. Table \ref{tab:opt_params} shows the optimization weights that we used.
\section{RESULTS}
The presented planning framework was validated on a variety of terrains. Each terrain is modelled by a height map using data from an Ouster OS0-128 LIDAR consisting of 128 vertical channels, 2048 scan points per channel, a 360-degree scan radius and a 10 Hz update rate. In this section we outline the process for building the map and testing the planner. A complete description of the sensor configuration is given in \cite{mishra22}.

\subsection{Height Map} \label{subsec:height_map}
\begin{figure}[t]
  \vspace{2mm}
\centering
  \includegraphics[width=0.97\columnwidth]{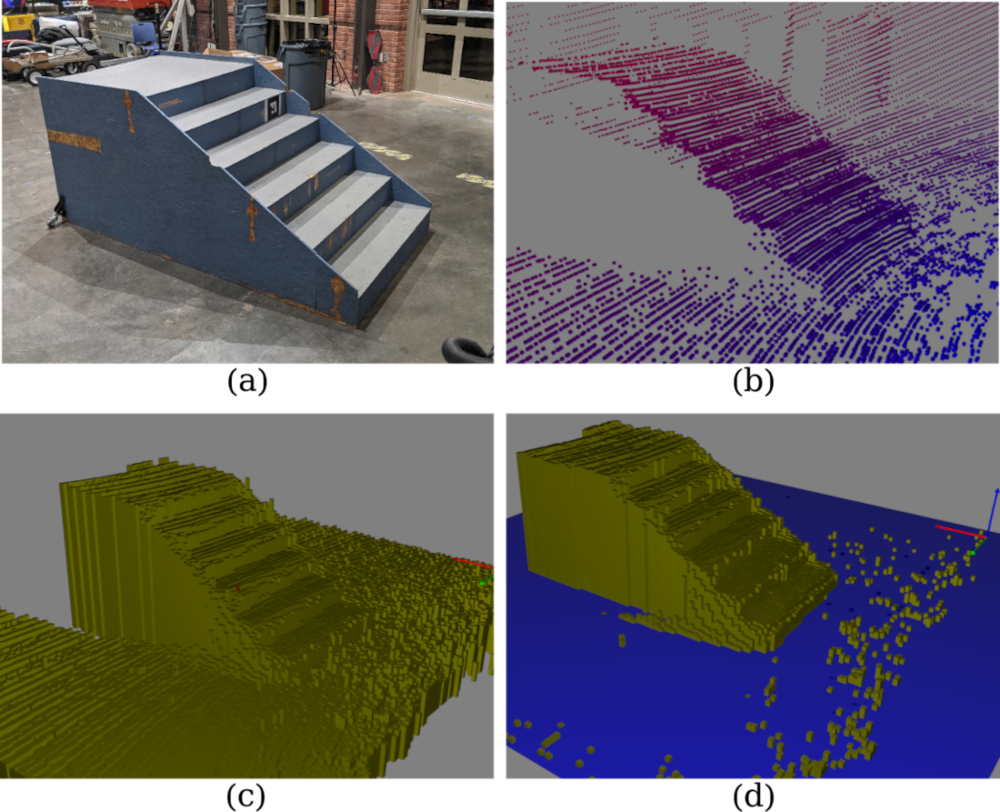}
  \caption{The height map is built using point clouds from an Ouster OS0-128 LIDAR. Points within the height map's domain are then assigned a cell, merged with previous data, then filtered to estimate the ground plan and remove outliers.}
  \label{fig:height_map}
  \vspace{-2mm}
\end{figure}

The pipeline for building the height map is shown in Fig. \ref{fig:height_map}. The map is characterized by a nominal $xy$ center position, a discritezation $\Delta_h$ and a grid width $w$. In order to refine the height map quality a number of simple filters are used following the approaches in \cite{kleiner2007real, marion2016}. The first filter estimates the ground plane height $h_G$ as the mean of all heights between the $2^{nd}$ and $6^{th}$ height percentiles. Heights below this are ignored to remove outliers.
\begin{equation}
    h_{G} = \mu(z_2 : z_6)
\end{equation}
The next filter clears all data for grid cells at a height below $h_G + \epsilon$ as these cells are likely part of the ground plane. Our calibration of the Ouster OS0-128 models point measurements as having Gaussian noise with a standard deviation of $\sigma=$1.8cm. Using this we set the ground height threshold to $\epsilon = 2\sigma=$3.6cm. Cells in which all neighbors are ground planes cells are also considered part of the ground plane. Finally, cells significantly higher than all 8 neighbors are assumed to be noisy outliers and reset to the local average height.

\subsection{Planner Performance}

\begin{figure}[t]
\centering
  \includegraphics[width=\columnwidth]{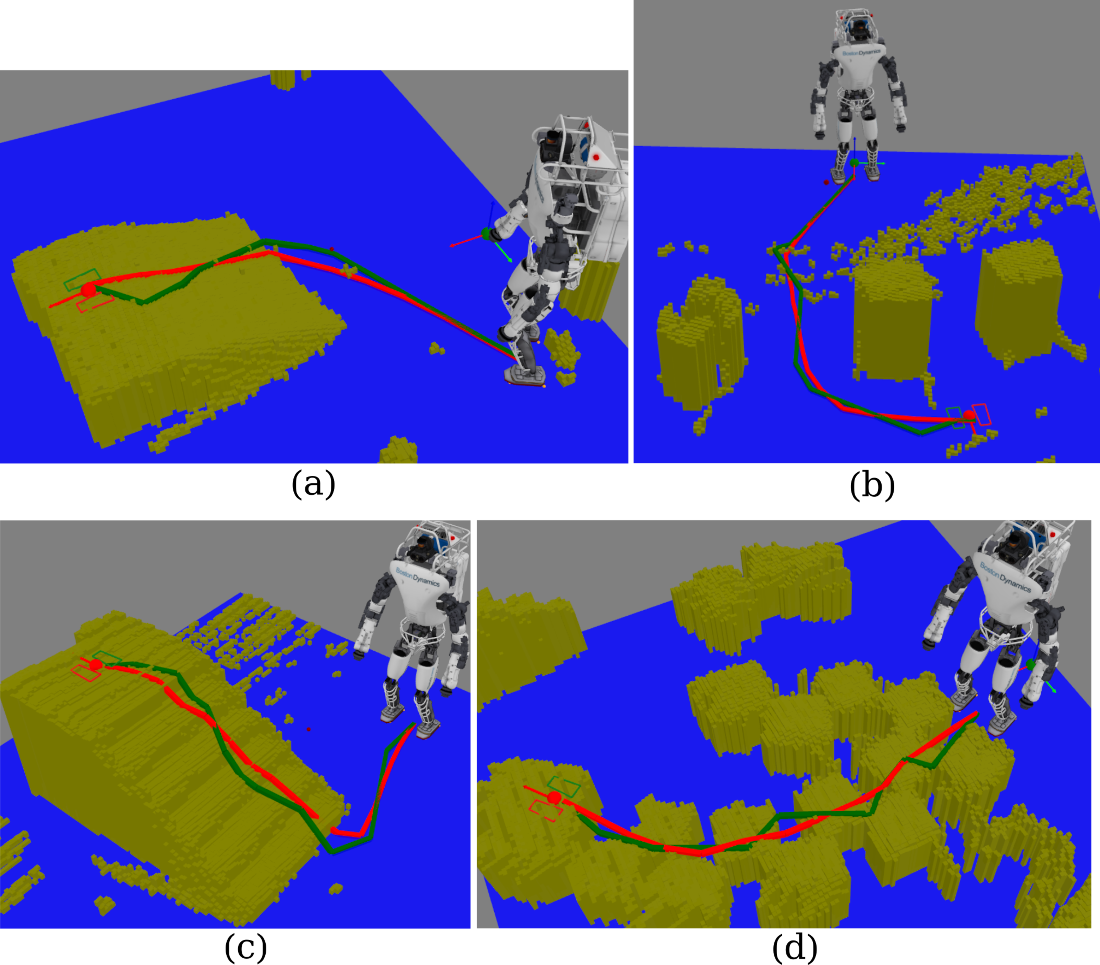}
  \caption{Simulation results for various rough terrains using real sensor data. The green path shows the initial path generated by A* and the red is the optimized path.}
  \label{fig:sim_plans}
  \vspace{-3mm}
\end{figure}

Fig. \ref{fig:sim_plans} shows the planner's performance on a variety of terrains. We focused on terrains that require accurate traversability modelling (a,c,d), collision detection (b) and include inclines (a,c). These maps were generated using the robot's LIDAR sensor and tested offline in order to aid the development and tuning process. Table \ref{tab:plan_stats} shows the resulting data from running on an Intel Core i9-9980HK at 2.4GHz.

\begin{table}
  \vspace{2mm}
\caption{Planner Performance on Test Terrains}

\begin{center}
\begin{tabular}{|c c c c c|} 
 \hline
 & \stack{A*}{Time (s)} & \stack{Optimizer}{Time (s)} & \stack{Path}{Length (m)} & \stack{Expanded}{Nodes} \\ [0.2ex] 
 \hline
 % Ramp
 Fig. \ref{fig:sim_plans}a & 0.64 & 1.42 & 3.93 & 792 \\ 

 % Obstacles
 Fig. \ref{fig:sim_plans}b & 0.45 & 1.46 & 5.52 & 355 \\

 % Stairs
 Fig. \ref{fig:sim_plans}c & 0.73 & 1.38 & 4.30 & 1079 \\ 

 % Stepping Stones
 Fig. \ref{fig:sim_plans}d & 0.51 & 1.36 & 3.89 & 268 \\

 % Obstacle Course
 Fig. \ref{fig:intro_figure} & 1.35 & 2.91 & 9.22 & 1137 \\

 \hline
\end{tabular}
  \vspace{-3mm}
\end{center}

\label{tab:plan_stats}
\end{table}

For the terrains in Fig. \ref{fig:sim_plans} the planner finds a solution on the order of 2s. Combined with 2s to accumulate sufficient data for the height map this results in approximately 4s end-to-end plan time and is thus capable of planning at operational speeds. Additionally, once a nominal plan is generated it can be subsequently re-validated to check for collisions and traversability at a much faster rate of approximately $n t_i$ where $n$ is the number of waypoints and $t_i=1.2$ms is the nominal time per A* iteration. The runtime could be improved by moving both the RANSAC and least-squares surface normal computation from the CPU to the GPU, as these are highly parallel calculations. On the CPU, the surface normal computation takes $15 \pm 3$\% of an A* iteration, depending on the amount of flat ground.

The ramp and stair terrains (Fig. \ref{fig:sim_plans}a,c) primarily test the contour cost term, which is demonstrated by the robot first walking to the base of the incline before walking up. However, increasing the contour weight too much can begin affecting plans with small contour changes where the cost would otherwise be negligible, such as the stepping stones terrain (Fig. \ref{fig:sim_plans}d). We tried adding an additional cost term which discounted the traversability of cells just above the ground plane, the idea being to mitigate the heuristic cost steering towards the goal while finding an ideal position to ``depart'' from the ground plane. However for cases such as the ramp where there is a gradual slope near the ground plane the heuristic cost could be outweighed, resulting in excessively long path lengths.

The traversability cost is primarily demonstrated in the stepping stones dataset (Fig. \ref{fig:sim_plans}d) by driving the path towards the center of the steps. It also contributes to avoiding the edge of the ramp and stairs while ascending, however, it will not inherently penalize small path segments near the edge based on the size of the preview window $p$. But in combination with the contour cost, such paths that shift edge-to-middle are accordingly penalized.

% Key reliability issue is identifying and cutting corners too close. Mitigate this by penalizing the roll cost and adding a term in for trusting ground cells. But this can through off the plan for the ramp when there is a smooth transition. Semantic labelling of waypoints could be crucial for this.

% Another issue is centering on the cinders blocks. But the footstep planner has a window of ~25cm within the body path and therefore it doesn't need to be perfect.

% Reasoning about individual contacts is great but doesn't seem feasible with the given resolution of the heightmap. It might be easier for quadrupeds as point-feet can be easier used on non-planar terrain.

\subsection{Results on DRC Atlas Robot}

\begin{figure}[t]
  \vspace{2mm}
\centering
  \includegraphics[width=0.95\columnwidth]{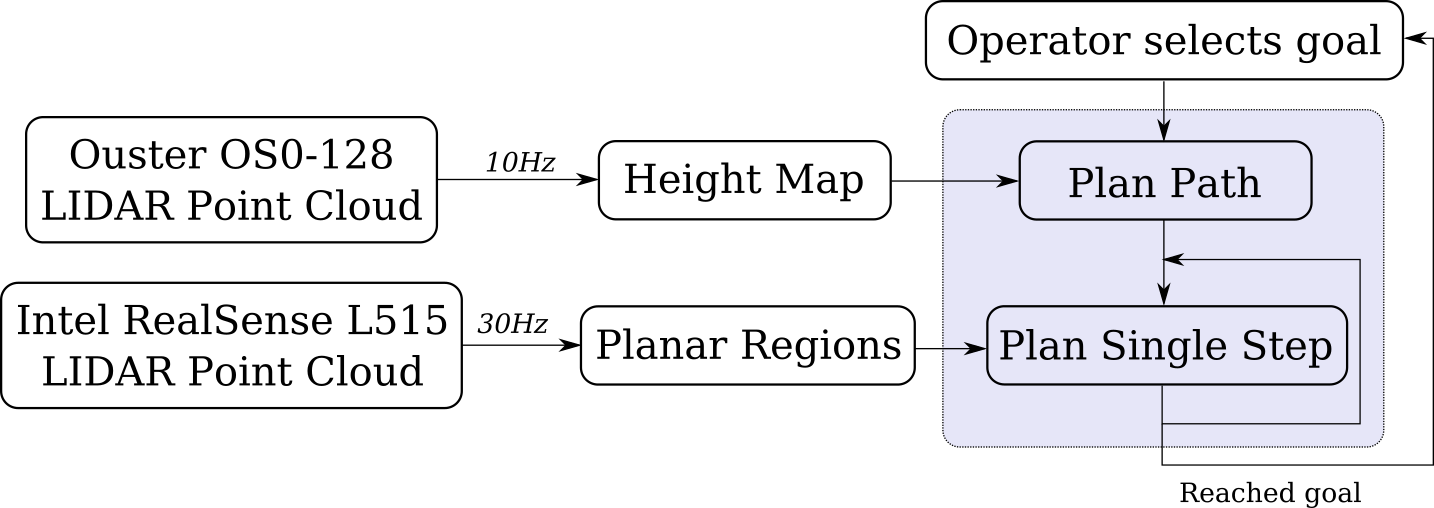}
  \caption{Integrated operator-directed navigation in which the initial path is planned over the height map and individual steps are subsequently planned using planar regions.}
  \label{fig:nav_framework}
  \vspace{-3mm}
\end{figure}

We integrated the presented path planner with the IHMC balance controller \cite{IHMCController} and footstep planner \cite{IHMCFootstepPlanner} to create a basic navigation framework, shown in Fig. \ref{fig:nav_framework}. The operator first selects a goal pose, which prompts the planner to compute a nominal body path. The robot then follows this path by iteratively planning and executing single steps. An Intel RealSense L515 LIDAR mounted on the pelvis generates high-resolution point clouds of the terrain immediately in front of the robot ($<$1m). An algorithm fits planar regions to these point clouds \cite{mishra2021gpu} which are used by a footstep planner to compute the best upcoming foothold. We limit footstep planning to a single step based on the limited field of view.

\begin{figure}[t]
  \vspace{2mm}
\centering
  \includegraphics[width=\columnwidth]{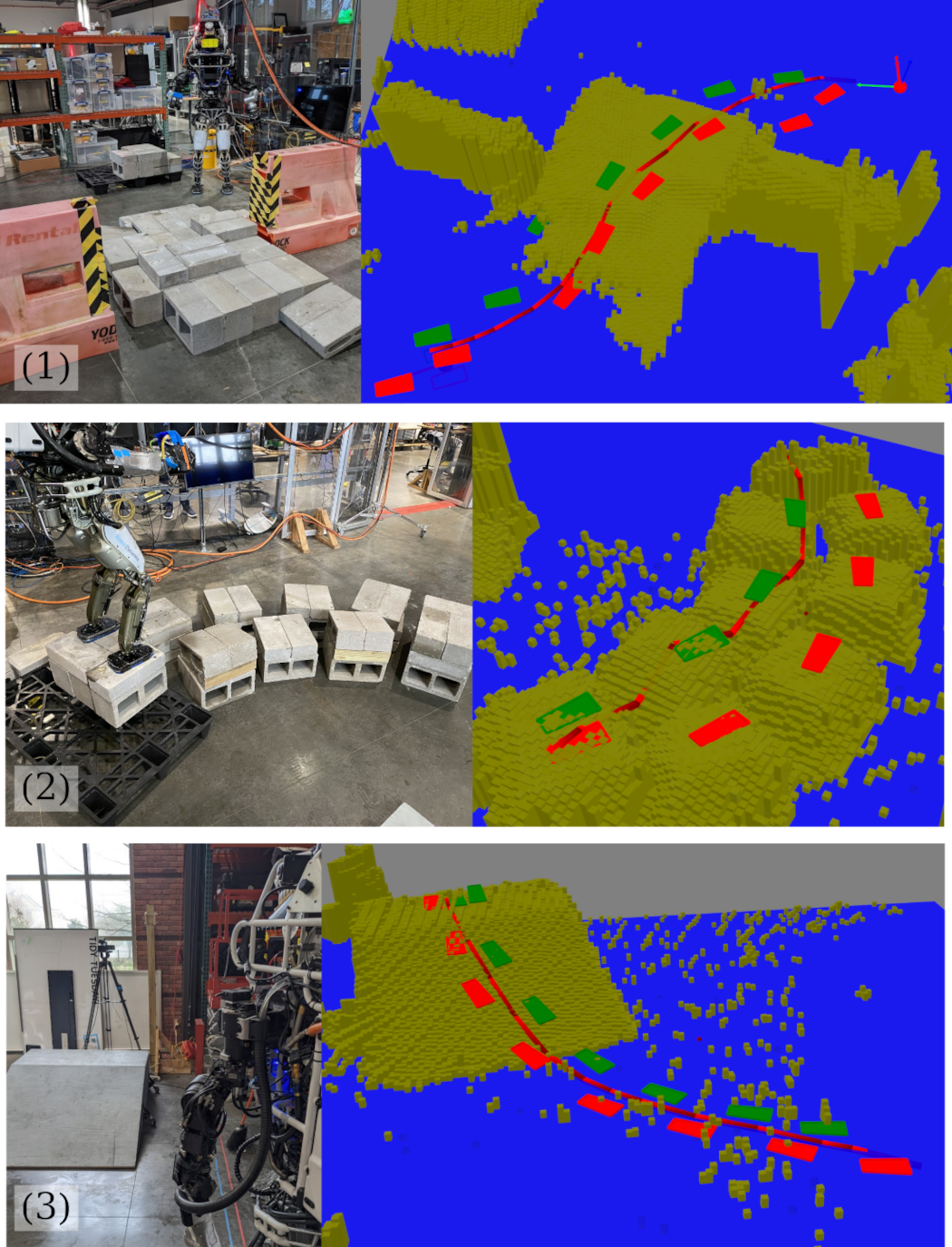}
  \caption{Results integrating the path planner with a balance controller and footstep planner on the Atlas robot. The optimized path is shown in red and the green/red steps are planned and executed as the robot walks using a short-horizon planar region map and single iteration A* footstep search.}
  \label{fig:robot_plans}
  \vspace{-4mm}
\end{figure}

The path and executed footsteps for the tested terrains are shown in Fig. \ref{fig:robot_plans}. The cinder block terrain (Fig. \ref{fig:robot_plans}-1) primarily tests the obstacle clearance and contour costs. The overlap between some of the executed steps and height map demonstrates how sensor noise and state estimator drift can be practical limitations for planning contacts at this range. For the stepping stones terrain (Fig. \ref{fig:robot_plans}-2), the path tends to be slightly on the inside edge of the center of the stepping stones but the robot still plans footholds which straddle either side. This offset is due to the contour cost being activated by slight height variation in the stepping stones. However, as the path is only a heuristic for the footstep planner, it is robust to this level of variation.

\section{CONCLUSION AND FUTURE WORK}

In this work we have presented a humanoid path planner designed to compute robust and natural paths. The goal of this work is to have a path planner which can enable a footstep planner and balance controller to traverse significant stretches of rough terrain, which is demonstrated over various terrains on hardware.

A common pitfall of A* search is the ``cul-de-sac problem,'' where a planner spends significant time exhausting an invalid route in a locally greedy direction toward the goal before moving on. This is seen to some extent in the stairs terrain, where the planner spends a significant percentage of the time searching the ground plane near the goal before finding a route up the stairs. Performing an initial pass with a simplified set of metrics could be used to improve the A* heuristic and expedite such plans. For the plan durations presented here, we did not find such improvements a high priority. However it could be the case for certain terrains this problem is exacerbated. We will take a first step in investigating this by analyzing the planner's runtime complexity.

We also intend to investigate using semantic labels to assist the A* plan. One way this could be combined with the given framework is having waypoints associated with semantic labels which could in turn be used as a heuristic for the path planner.

\subsection{Source Code}
The source code for the height map and path planner can be found on our GitHub page at \url{https://github.com/ihmcrobotics/ihmc-open-robotics-software}, Java packages \texttt{us.ihmc.\{robotics.heightMap, footstepPlanning.bodyPath\}}.
\printbibliography
\balance      %ADDED 20220224 HES
\end{document}